\documentclass[10pt,a4paper]{article}
\usepackage[utf8]{inputenc}
\usepackage{amsmath}
\usepackage{amsfonts}
\usepackage{amssymb}
\usepackage{graphicx}%
\usepackage{multirow}%
\usepackage{amsmath,amssymb,amsfonts}%
\usepackage{amsthm}%
\usepackage{mathrsfs}%
\usepackage[figuresright]{rotating}%
\usepackage[title]{appendix}%
\usepackage{xcolor}%
\usepackage{textcomp}%
\usepackage{manyfoot}%
\usepackage{booktabs}%
\usepackage{algorithm}%
\usepackage{algorithmicx}%
\usepackage{algpseudocode}%
\usepackage{program}%
\usepackage{listings}%
\usepackage{url}
\usepackage{geometry}

\newcommand{\err}[1]{\color {gray}\footnotesize $\pm #1$}

\newcommand{\sig}[1]{}
\usepackage{xcolor}

\newcommand{\na}{-}
\usepackage{graphicx}
\usepackage{authblk}
\usepackage{natbib}

\begin{document}

\title{
Uncertainty-aware Pseudo-label Selection for Positive-Unlabeled Learning
}

\author[1,2,3]{Emilio Dorigatti}
\author[1,3,4]{Jann Goschenhofer}
\author[2,5]{Benjamin Schubert}
\author[1]{Mina Rezaei}
\author[1,3,4]{Bernd Bischl}

\affil[1]{\small Department of Statistics, Ludwig-Maximilians-Universit\"at M\"unchen, M\"unchen 80539, Germany}
\affil[2]{\small Institute of Computational Biology, Helmholtz Zentrum M\"unchen—German Research Center for Environmental Health, Neuherberg 85764, Germany}
\affil[3]{\small Munich Center for Machine Learning, M\"unchen, Germany}
\affil[4]{\small Fraunhofer Institute for Integrated Circuits IIS, Erlangen 91058, Germany}
\affil[5]{\small Department of Mathematics, Technical University of Munich, Garching bei M\"unchen 85748, Germany}

\renewcommand\Authands{ and }

\maketitle

\textbf{Abstract:}
Positive-unlabeled learning (PUL) aims at learning a binary classifier from only positive and unlabeled training data.
Even though real-world applications often involve imbalanced datasets where the majority of examples belong to one class, most contemporary approaches to PUL do not investigate performance in this setting, thus severely limiting their applicability in practice.
In this work, we thus propose to tackle the issues of imbalanced datasets and model calibration in a PUL setting through an uncertainty-aware pseudo-labeling procedure (\emph{PUUPL}): by boosting the signal from the minority class, pseudo-labeling expands the labeled dataset with new samples from the unlabeled set, while explicit uncertainty quantification prevents the emergence of harmful confirmation bias leading to increased predictive performance.
Within a series of experiments, PUUPL yields substantial performance gains in highly imbalanced settings while also showing strong performance in balanced PU scenarios across recent baselines.
We furthermore provide ablations and sensitivity analyses to shed light on PUUPL's several ingredients. 
Finally, a real-world application with an imbalanced dataset confirms the advantage of our approach.


\section{Introduction}
\label{introduction}
Many real-world applications involve positive-unlabeled (PU) datasets \cite{armenian1974distribution,gligorijevic2021structure,mass_spec} in which only a few samples are labeled positive while the majority is unlabeled.
PU learning (PUL) aims to learn a binary classifier in this challenging setting without any labeled negative examples, thus reducing the need for manual annotation and enabling entirely new applications where negative examples are costly or impossible to obtain~\citep{Kiryo2017}.
Learning from PU data can reduce development costs in many deep learning applications that otherwise require costly annotations from experts or expensive experimental procedures such as medical image diagnosis~\cite{armenian1974distribution} and protein function prediction~\cite{gligorijevic2021structure}.
PUL can even enable applications in settings where the measurement technology itself can not detect negative examples~\cite{mass_spec}.

Many PUL applications share another intrinsic difficulty: class imbalance.
Imbalanced settings arise when most samples in a dataset belong to the same class, and frequently the most interesting class happens to be the minority.
In PUL, class imbalance refers specifically to a low class prior $\pi:=p(y=1)$ implying that the majority of the unlabeled samples are negatives.
While this problem can be tackled in traditional (semi-)supervised learning by re-weighting the loss to increase the penalty of mis-classification of the minority class, a similar approach was introduced in PUL with some additional care in handling the unlabeled data points~\citep{imbnnpu}. 
However, the issue remains in general under-studied in the literature and recent developments such as Self-PU \citep{selfpu} are solely targeted at balanced scenarios.

\begin{figure*}[t]
    \centering
    \includegraphics[width=0.8\linewidth]{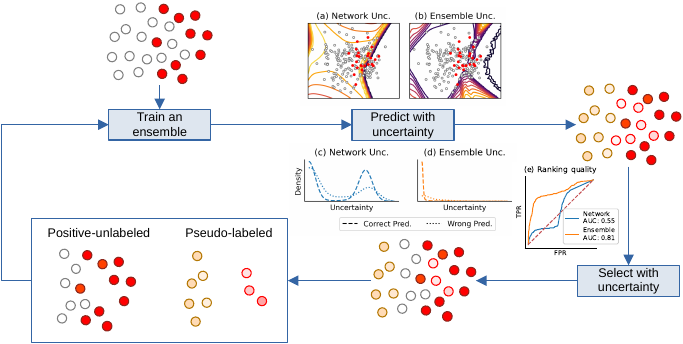}
    \caption{
    \small
    \emph{PUUPL} is a pseudo-labeling framework for PU learning that uses the epistemic uncertainty of an ensemble to select confident examples to pseudo-label.
    The ensemble can be trained with any PU loss for PU data while minimizing the cross-entropy loss on the previously assigned pseudo-labels.
    In a toy example, a single network is not very confident on most of the unlabeled data (a), resulting in many high-confidence incorrect predictions and many low-confidence correct ones (c).
    The epistemic uncertainty of an ensemble is, on the other hand, very low on most of the unlabeled data (b), resulting in most correct predictions having low uncertainty and most incorrect predictions having high uncertainty (d).
    Thus, the estimated uncertainty by ensemble can be used more reliably to rank predictions and select correct ones (e).
    Re-training the model with an increased number of labeled samples will result in a slightly more accurate model, than can be used to predict new pseudo-labels, which will further improve the model's performance, etc.
    }
    \label{fig:flowchart}
\end{figure*}

Motivated by this, we propose to tackle imbalancedness in PUL via pseudo-labeling \citep{Lee2013}, an iterative procedure that augments the labeled dataset with new samples from the unlabeled set, thus boosting the weak signal from the minority class.
To prevent the emergence of harmful confirmation bias in this procedure, we propose to assign pseudo-labels based on likelihood-free uncertainty quantification via model ensembling \citep{deepens}.
By using soft targets we avoid artificially inflating the confidence of pseudo-labels and preserve the calibration signal for the ensemble in later training iterations, thus eventually obtaining a predictor that is both calibrated and well-performing.
Another advantage of pseudo-labeling is that it allows the model to harness the power of self-training without requiring modality-specific augmentations such as MixUp~\citep{mixup} that restrict most contemporary PUL methods~\citep{vpu,apu,punce,selfpu,distpu} to image data only.

To summarize, our contributions are:
\begin{enumerate}
\item We introduce \emph{PUUPL} (Positive UnlabeledUncertainty aware Pseudo-Labeling), a novel framework that successfully overcomes the issue of imbalanced data distribution in PUL in a data-modality-agnostic framework while retaining competitive performance on balanced datasets.
\item We evaluate our methods on a wide range of benchmarks and PU datasets, achieving state-of-the-art results in self-training for PUL both with and without knowing the positive class prior $\pi$. 
Our results show that PUUPL is applicable to different data modalities such as images and text, can use any risk estimator for PUL and improve thereupon, and is robust to prior misspecification and class imbalance.
\item We apply PUUPL to a real-world healthcare dataset, confirming the advantage of PUUPL compared to other PUL methods as well as previous domain-specific state-of-the-art approaches.
\end{enumerate}
These results demonstrate that our framework is highly reliable, extensible, and applicable in a variety of real-world scenarios.

\section{Related work}

PUL was introduced as a variant of binary classification \cite{Liu2003} and is related to one-class learning \cite{ruff2018deep,li2010positive}, multi-positive learning~\cite{xu2017multi}, multi-task learning~\cite{KajiYS18}, and semi-supervised learning~\cite{Chapelle2009}.
Current existing methods for PUL can be divided into three branches: two-step techniques, class prior incorporation, and biased PUL \cite{bekker2020}.
In this work, we apply pseudo-labeling with biased PUL -- also coined as re-weighting methods -- and refer to Bekker et al. \cite{bekker2020} for a comprehensive overview of the field. 
In this context, Du Plessis et al. \cite{plessis2014} introduced the unbiased risk estimator uPU. 
Kiryo et al. \cite{Kiryo2017} showed that this loss function is prone to overfitting in deep learning contexts, as it lacks a lower bound, and proposed the non-negative risk estimator nnPU \cite{Kiryo2017} as a remedy. 
Follow-up work on loss functions for PUL has focused on robustness w.r.t. biases in the sampling process~\cite{Kato2019,Hsieh2019,Luo2021} and handling of imbalanced datasets \cite{imbnnpu}.
Further research in PUL focuses on estimating the class prior directly during training~\cite{vpu,gargMEP,Hu2021PredictiveAL} or exploiting its knowledge to further improve the training process~\cite{selfpu,punce,distpu,apu}.

Pseudo-labeling~\citep{Lee2013} is a popular approach for semi-supervised learning when several unlabeled examples are available.
In pseudo-labeling, the model leverages its own predictions on unlabeled data as training targets to enable iterative semi-supervised model training.
However, erroneously selected pseudo-labels can amplify errors during training, potentially leading to model degradation over time.
This \textit{confirmation bias} is grounded in poor model calibration which distorts the signal for the pseudo label selection \cite{VanEngelen2020}, and is exacerbated in a deep learning setting as deep neural networks are prone to over-confident predictions unless trained appropriately~\cite{Gawlikowski2021ASO}. 
A variety of approaches were proposed for semi-supervised classification settings to mitigate this problem~\citep{Iscen2019,Shi2018,Arazo2020PseudoLabelingAC,Tanaka2018,Rizve2021,Beluch2018}: the commonality of these works is the explicit consideration of model uncertainty to improve pseudo-label selection, which motivates its application in the context of PUL. 

A first attempt to combine pseudo-labeling with PUL was made with Self-PU~\cite{selfpu}, where self-paced learning, a confidence-weighting scheme based on the model predictions, and a teacher-student distillation approach are combined.
With \textit{PUUPL}, we propose an alternative pseudo-labeling strategy for PUL that performs better in a simpler and more principled way using implicitly well-calibrated models to improve the pseudo-label selection. 
Moreover, uncertainty awareness allows \textit{PUUPL} to work well in unbalanced data environments where Self-PU breaks down.
To the best of our knowledge, we are the first to introduce an uncertainty-aware pseudo-labeling paradigm to PUL.
Although our method shares the same motivation as that from Rizve et al. \cite{Rizve2021} for semi-supervised classification with both positive and negative training samples, we differ in several important aspects dictated by the PUL setting: (1) we specifically target PU data with a PU loss, (2) we quantify uncertainty with an ensemble instead of Monte Carlo dropout, (3) we use epistemic uncertainty instead of the predicted class probabilities for the selection, (4) we do not use temperature scaling and (5) use soft labels.

\section{Method}
PUUPL (Algorithm~\ref{algo:pl}) separates the training set $X^{tr}$ into the sets $P$, $U$, and $L$, which contain the initial positives, the currently unlabeled, and the pseudo-labeled samples, respectively. 
The set $L$ is initially empty. 
At each pseudo-labeling iteration, we first train our model using all samples in $P$, $U$, and $L$ until some convergence condition is met (Section~\ref{sec:loss}). 
Then, model predictions over the samples in $U$ are ranked w.r.t. their predictive uncertainty (Section \ref{sec:unc}), and samples with the most confident score are assigned the predicted label and moved into the set $L$ (Section \ref{sec:pl}). 
Similarly, model predictions are derived for the samples in $L$, and the most uncertain samples are moved back to the unlabeled set $U$ (Section~\ref{sec:unlab}). 
Next, the model is re-initialized to the same initial weights, and the next pseudo-labeling iteration starts.

\subsection{Notation}\label{sec:notation}
Consider input samples $X$ with label $y$ and superscripts ${\cdot}^{tr}$, ${\cdot}^{va}$ and ${\cdot}^{te}$ for training, validation, and test data, respectively. 
The initial training labels $y^{tr}$ are set to one for all samples in $P$ and zero for all others in $U$.
We group the indices of original positives, unlabeled, and pseudo-labeled samples in $X^{tr}$ into the sets $P$, $U$, and $L$, respectively, where $y_i=1$ for $i\in P$, $y_i=0$ for $i\in U$, and $y_i$ is the assigned pseudo-label for $i\in U$.
Our proposed model is an ensemble of $K$ deep neural networks whose random initial weights are collectively denoted as $\theta^0$. 
The predictions of the $k$-th network for sample $i$ are indicated with $\hat{p}_{ik}=\sigma(\hat{f}_{ik})$, with $\sigma(\cdot)$ as the logistic function and $\hat{f}_{ik}$ as the predicted logits.
The logits and predictions for a sample averaged across the networks in the ensemble are denoted by $\hat{f}_i$ and $\hat{p}_i$, respectively.
We subscript data and predictions with $i$ to index individual samples, and use an index set in the subscript to index all samples in the set (e.g., $X^{tr}_U=\{x^{tr}_i \vert i\in U\}$ denotes the features of all unlabeled samples).
We denote the total, epistemic, and aleatoric uncertainty of sample $i$ as $\hat{u}^t_i$, $\hat{u}^e_i$, and $\hat{u}^a_i$, respectively.

\begin{algorithm}[t]
\caption{The PUUPL Training Procedure}\label{alg:cap}
\label{algo:pl}
\small
\hspace*{\algorithmicindent} \textbf{Hyperparameters:}\begin{itemize}
\setlength\itemsep{0.0em}
\item Loss mixing coefficient $\lambda$
\item Number $K$ of networks in the ensemble
\item Maximum number $T$ of pseudo-labels to assign at each round
\item Maximum uncertainty threshold $t_l$ to assign pseudo-labels
\item Minimum uncertainty threshold $t_u$ to remove pseudo-labels
\end{itemize}
\hspace*{\algorithmicindent} \textbf{Input:} Train and validation $X^{tr}, y^{tr}, X^{va}, y^{va}$
\begin{algorithmic}[1]
\State $P\gets$ indices of positive samples in $X^{tr}$
\State $U\gets$ indices of unlabeled samples in $X^{tr}$
\State $L\gets\emptyset$
\State $\theta^0 \gets $ Random weight initialization 
\While{not converged}
\State Initialize model weights to $\theta^0$
\State Train an ensemble of $K$ networks on $X^{tr},y^{tr}$ 
\State Update $\theta^*$ if performance on $X^{va},y^{va}$ improved
\State $\hat{f} \gets$ ensemble predictions for $X^{tr}$
\State Compute epistemic uncertainty via Eq. \ref{eq:unc_e}
\State $L^{\text{new}}\gets$ Examples to pseudo-label via Eq.~\ref{eq:lnew} 
\State $U^{\text{new}}\gets$ Examples to pseudo-unlabel 
\State $L \gets L\cup L^{\text{new}}_b \setminus U^{\text{new}}$
\State $U\gets U \setminus L^{\text{new}}_b \cup U^{\text{new}}$
\State $y_{L^{\text{new}}}\gets\hat{p}_{L^{\text{new}}}$
\State $y_{U^{\text{new}}}\gets 0$
\EndWhile
\end{algorithmic}
\end{algorithm}


\subsection{Loss function} \label{sec:loss}
We train our proposed model with a loss function $\mathcal{L}$ that is a convex combination of a loss $\mathcal{L}_{PU}$ for the samples in the positive and unlabeled set ($P\cup U$) and a loss $\mathcal{L}_L$ for the samples in the pseudo-labeled set ($L$):
\begin{equation}
\label{eq:combined_loss}
\mathcal{L} = \lambda\cdot\mathcal{L}_L + (1 - \lambda)\cdot\mathcal{L}_{PU}
\end{equation}
with $\lambda\in(0,1)$.
The loss $\mathcal{L}_L$ is the binary cross-entropy computed w.r.t. the assigned pseudo-labels.
Our method is agnostic to the specific PU loss $\mathcal{L}_{PU}$ used, allowing PUUPL to be easily adapted to provide further performance increase in other scenarios for which a different PU loss might be more appropriate, e.g., when a set of biased negative samples is available \cite{Hsieh2019}, when coping with a selection bias in the positive examples \cite{Kato2019} or an imbalanced class distribution~\cite{imbnnpu} (see experiments).
For the standard setting of imbalanced PUL, we use the imbnnPU loss~\citep{imbnnpu}:
\begin{equation}
\label{eq:imbnnpu_loss}
\mathcal{L}_{PU}= \pi^\prime\ell(P, 1) + \max\left\{0,
\frac{1-\pi^\prime}{1-\pi}\ell(U, -1) 
-\frac{(1-\pi^\prime)\pi}{1-\pi}
\ell(P, -1) \right\}
\end{equation}
where $\pi=p(y=1)$ is the prior probability that a sample is positive, $\pi^\prime$ the desired oversampled probability that we fix to $1/2$,
and $\ell(S, y)$ the eWe do not run bibtex in the auto-TeXing procedure. If you use bibtex, you must compile the .bbl file on your computer then include that in your uploadedxpected sigmoid loss of samples in the set $S$ with label $y$:
\begin{equation}
\ell(S, y) = \frac{1}{\vert S \vert}\sum_{i\in S}\frac{1}{1+\exp(y\cdot\hat{p}_i)}
\label{eq:sigmoid}
\end{equation}

Similarly, we use use the non-negative correction $nnPU$ of the PU loss~\cite{Kiryo2017} for the standard, balanced PU setting:
\begin{equation}
\label{eq:nnpu_loss}
\mathcal{L}_{PU}= \pi\cdot\ell(P, 1) + \max\left\{0, \ell(U, -1) - \pi\cdot\ell(P, -1) \right\}
\end{equation}
where $\pi:=p(y=1)$ is the prior probability that a sample is positive
and $\ell(S, y)$ follows equation \ref{eq:sigmoid}.

While $\pi$ can be estimated from PU data \cite{marthinus2016}, in our experimental results we treat $\pi$ as a hyperparameter and optimize it without requiring negatively labeled samples via a PU validation set~\cite{Menon2015LearningFC}.


\subsection{Model uncertainty} \label{sec:unc}
We utilize a deep ensemble with $K$ networks with the same architecture, each trained on the full training dataset \cite{deepens}, to quantify the predictive uncertainty. 
Given the predictions $\hat{p}_{i1},\ldots,\hat{p}_{iK}$ for a sample $x_i$, we associate three types of uncertainties to $x_i$'s predictions \cite{Hllermeier2021AleatoricAE}: the aleatoric uncertainty as the mean of the entropy of the predictions (Eq.~\ref{eq:unc_a}), the total uncertainty as the entropy of the mean prediction (Eq.~\ref{eq:unc_t}), and the epistemic uncertainty formulated as the difference between the two (Eq.~\ref{eq:unc_e}).
\begin{align}
\label{eq:unc_a}
\hat{u}^a_i&=-\frac{1}{K}\sum_{k=1}^K\left[
\hat{p}_{ik}\log\hat{p}_{ik}+
(1-\hat{p}_{ik})\log(1-\hat{p}_{ik})
\right] \\
\label{eq:unc_t}
\hat{u}^t_i&=-\hat{p}_i\log\hat{p}_i-(1-\hat{p}_i)\log(1-\hat{p}_i) \\
\label{eq:unc_e}
\hat{u}^e_i&=\hat{u}^t_i-\hat{u}^a_i
\end{align}
where $\hat{p}_i=\sum_{k=1}^K\hat{p}_{ik}/K$.
Epistemic uncertainty corresponds to the mutual information between the parameters of the model and the true label of the sample.
Low epistemic uncertainty thus means that the model parameters would not change significantly if trained on the true label, suggesting that the prediction is indeed correct.
The cumulative effect of many correct pseudo-labels added over time, however, provides a strong enough training signal to push the model towards better-performing parameters, as we show in the experimental results.


\subsection{Pseudo-labeling} \label{sec:pl}
The estimated epistemic uncertainty (Eq.~\ref{eq:unc_e}) is used to rank and select unlabeled examples for pseudo-labeling.
Let $\rho(i)$ denote the rank of sample $i$. 
Then, the set $L^{\text{new}}$ of newly pseudo-labeled samples is formed by taking the $T$ samples with lowest uncertainty from $U$, ensuring that it is lower than a threshold $t_l$:
\begin{equation}
\label{eq:lnew}
L^{\text{new}}
= \left\{
i\in U \vert \rho(i) \leq T \land u^e_i \leq t_l
\right\} \\
\end{equation}
Previous works on semi-supervised classification have shown that balancing the pseudo-label selection between the two classes -- i.e., ensuring that the ratio of newly labeled positives and negatives is close to a given target ratio $r$ -- is beneficial~\cite{Rizve2021}.
In this case, the set $L^{\text{new}}$ is partitioned according to the model's predictions into a set
$L^{\text{new}}_+$ of predicted positives and $L^{\text{new}}_-$ of predicted negatives, and the most uncertain samples in the larger set are discarded to reach the desired ratio $r$, which we fix to $1$.
We then assign soft pseudo-labels, i.e., the average prediction in the open interval $(0,1)$, to these samples:
\begin{equation}
y_i = \hat{p}_i \quad \forall i \in L^{\text{new}}_-\cup L^{\text{new}}_+
\label{eq:yi_new}
\end{equation}
As discussed previously, low epistemic uncertainty signals likely correct predictions.
Using such predictions as a target in the loss $\mathcal{L}_{L}$ provides a stronger, more explicit learning signal to the model, resulting in a larger decrease in risk compared to using the same example as unlabeled in $\mathcal{L}_{PU}$.
At the same time, soft pseudo-labels provide an additional signal regarding the estimated aleatoric uncertainty of samples. 
Furthermore, they help reduce overfitting and the emergence of confirmation bias in case the assigned pseudo-label is wrong by acting as dynamically-smoothed labels~\cite{labelsmooth,Arazo2020PseudoLabelingAC}.

\subsection{Pseudo-unlabeling} \label{sec:unlab}
Similar to the way that low uncertainty on an unlabeled example indicates that the prediction can be trusted, high uncertainty on a pseudo-labeled example indicates that the assigned pseudo-label might not be correct after all. 
To avoid training on such possibly incorrect pseudo-labels, we move the pseudo-labeled examples with uncertainty above a threshold $t_u$ back into the unlabeled set:
\begin{align}
U^{\text{new}}=\left\{
i\in L \vert \hat{u}^e_i \geq t_u
\right\} \\
\label{eq:unlab}
y_i = 0 \quad \forall i\in U^{\text{new}}
\end{align}

\section{Experiments}
To empirically compare our proposed framework to existing state-of-the-art losses and models, we followed standard protocols for PUL~\cite{Kiryo2017,Kato2019,vpu,selfpu}.
After presenting the main results, we empirically show the advantage of our framework in improving performance for both imbalanced and standard PU scenarios, being applicable to different data modalities, and using various losses for PU learning.
Finally, we provide further analyses of PUUPL including an investigation of its sensitivity with respect to pseudo-labeling hyperparameters. 
The source code of the method and all the experiments are available at \url{https://anonymous.4open.science/r/PUUPL-BE6E}.


\subsection{Experimental protocol}
\label{sec:protocol}

\subsubsection{Datasets} 
We evaluated our method in the standard setting of MNIST~\cite{deng2012mnist} and CIFAR-10~\cite{krizhevsky2009learning} datasets, as well as Fashion MNIST (F-MNIST)~\cite{xiao2017fashion}, CIFAR-100-20~\cite{krizhevsky2009learning} and IMDb~\cite{IMDB2011} to show the applicability to different data modalities.
Similar to previous studies~\cite{vpu,Kiryo2017,Kato2019,selfpu}, positives were defined as odd digits in MNIST and vehicles in CIFAR-10.
For F-MNIST we used trousers, coats, and sneakers as positives, and for the experiments on CIFAR-100-20, we defined those 10 out of the 20 superclasses as positives that correspond to living creatures (i.e., `aquatic mammal`, `fish`, `insects`, `large carnivores`, `large omnivores`, `medium-sized mammals`, `non-insect invertebrates`, `people`, `reptiles`, `small mammals`). 
The number of training samples is reported in Supplementary Table~\ref{tbl:dsets}.

For all datasets, we reserved a validation set of 5,000 samples and used all other samples for training, evaluating on the canonical test set.
To simulate an imbalanced setting, we downsampled the positives in the training and validation sets to obtain $\pi=0.1$ and labeled only 600 of them.
We also report results with 1,000 and 3,000 randomly chosen labeled training positives without downsampling as is common in the literature.
For the image datasets, we subtracted the mean pixel intensity in the training set and divided it by the standard deviation, while for IMDb we used pre-trained GloVe embeddings of size 200 on a corpus of six billion tokens.

\subsubsection{Network architectures}
To ensure a fair comparison with other works in PUL \cite{selfpu,vpu,Kiryo2017}, we used the same architectures on the same datasets, namely a 13-layer convolutional neural network (CNN) for the experiments on CIFAR-10 and CIFAR-100-20 (Table~\ref{tbl:c10net}) and a multi-layer perceptron (MLP) with four fully-connected hidden layers of 300 neurons each and ReLU activation for MNIST and F-MNIST.
For IMDb, we used a bidirectional LSTM network with a MLP head whose number of units was optimized as part of the hyperparameter search (Table~\ref{tbl:imdbnet}).

\subsubsection{Training}
We trained all models with the Adam optimizer \cite{Kingma2015AdamAM} with $\beta_1=0.9$ and $\beta_2=0.999$ and an exponential learning rate decay with $\gamma=0.99$, while learning rate, batch size, and weight decay were tuned together with the other pseudo-labeling hyperparameters using the ranges in Table~\ref{tbl:hpars} in the Supplements.
We provide experimental results using both a PU validation set, to provide a real-world performance estimate, as well as a fully-labeled (PN) validation set to compare against state-of-the-art PUL methods that used such a labeled validation set~\cite{selfpu,Hu2021PredictiveAL} and to showcase the \emph{potential} of our method.
When using a PU validation set, we used the AUROC between positive and unlabeled samples as tuning criterion, as previous work~\cite{Menon2015LearningFC,Jain2017RecoveringTC} has shown that higher AUROC on PU data directly translates to higher AUROC on fully labeled data.

\subsubsection{Evaluation}
We obtain the final results by training the model five times with random initialization and training-validation split while using the same canonical test set, reporting the highest test accuracy obtained both with a fully-labeled (PN) and a PU validation set.
Statistical significance was established with an unpaired $t$-test comparing the performance of PUUPL and the best performer amongst all other methods.

We compare PUUPL against VPU~\cite{vpu} and Self-PU~\cite{selfpu} using the same network architecture and data splits.
We consider the former as it does not require a known prior $\pi$ and can use a PU validation set, and the latter as the state-of-the-art self-training method for PUL even though it requires a positive-negative (PN) labeled validation set.
We additionally compare against a naive, uncertainty-unaware, pseudo-labeling baseline ``+PL'' that used the sigmoid outputs directly as a ranking measure for pseudo-labeling instead of the epistemic uncertainty, while still assigning soft pseudo-labels.

\subsection{Main Results}
\label{sec:best_results}

\begin{table*}
\resizebox{\textwidth}{!}{%
\centering
\begin{tabular}{ l l l l l l l }
\toprule
& &
\multicolumn{5}{c}{Dataset } \\
\cmidrule{3-7}
Valid. &
Method &
MNIST &
F-MN  &
C-100-20  &
CIF-10  & 
IMDb \\
\cmidrule(r){1-2}\cmidrule(lr){3-7}
\multirow{4}{*}{PN} & Self-PU~\cite{selfpu} & 94.44\err{0.12} & 90.99\err{0.47} & 50.00\err{0.0} & 63.97\err{3.97} & \phantom{00}\na{} \\
& imbnnPU~\cite{imbnnpu} & 
95.65\err{0.11} &
91.54\err{0.18} &
71.61\err{0.73} &
87.59\err{0.26} &
74.44\err{0.61} \\
& \quad +PL &
95.19\err{0.20} &
91.26\err{0.22} &
71.80\err{0.93} &
85.82\err{0.50} &
75.94\err{0.61} \\
& \quad +PUUPL &
\textbf{96.09\err{0.10}} &
\textbf{91.93\err{0.12}} &
\textbf{73.79\err{0.33}} &
\textbf{88.93\err{0.31}} &
\textbf{78.16\err{0.78}} \\
\cmidrule(r){1-2}\cmidrule(lr){3-7}
\multirow{4}{*}{PU} & VPU~\cite{vpu} & 80.87\err{3.24} & 89.30\err{0.98} & 70.01\err{0.96} & 86.41\err{0.78} & \phantom{00}\na{} \\
& imbnnPU~\cite{imbnnpu} &
\textbf{95.61\err{0.05}} &
89.88\err{0.51} &
67.13\err{1.26} &
87.61\err{0.25} &
74.32\err{0.58} \\
& \quad+PL  &
94.65\err{0.48} &
89.55\err{0.57} &
68.42\err{1.18} &
84.57\err{1.32} &
75.88\err{0.68} \\
& \quad+PUUPL  &
95.30\err{0.50} &
\textbf{91.86\err{0.09}} &
\textbf{72.80\err{0.38}} &
\textbf{87.97\err{0.38}} &
\textbf{77.78\err{0.86}} \\
\bottomrule \\
\end{tabular}
}
\caption{\small Average test accuracy and its standard error over five repetitions where model training was performed with an imbalanced dataset with $\pi=0.1$ and 600 labeled positives.
The row ``+PL`` refers to an uncertainty-unaware pseudo-labeling baseline, while ``+PUUPL`` refers to our uncertainty-aware solution.
The validation column refers to the use of a fully-labeled (PN) or PU validation set.
}
\label{tbl:imb_acc}
\end{table*}

\begin{table*}
\resizebox{\textwidth}{!}{%
\centering
\begin{tabular}{ l l l l l l l }
\toprule
& &
\multicolumn{5}{c}{Dataset } \\
\cmidrule{3-7}
Valid. &
Method &
MNIST &
F-MN  &
C-100-20  &
CIF-10  & 
IMDb \\
\cmidrule(r){1-2}\cmidrule(lr){3-7}
\multirow{4}{*}{PN} &
Self-PU~\cite{selfpu} & 95.64\err{0.13} & 91.55\err{0.18} & 75.41\err{0.44} & 90.56\err{0.09} & \phantom{00}\na{} \\
& nnPU~\cite{Kiryo2017} &
96.36\err{0.06} &
91.70\err{0.12} &
72.46\err{0.83} &
90.49\err{0.13} &
79.62\err{0.67}
\\
& \quad +PL             &
96.22\err{0.13} &
92.09\err{0.12} &
74.07\err{0.71} &
90.56\err{0.11} &
79.04\err{0.39} \\
& \quad +PUUPL &
\textbf{97.02\err{0.08}} &
\textbf{92.13\err{0.09}} &
\textbf{77.39\err{0.31}} &
\textbf{91.12\err{0.04}} &
\textbf{80.43\err{0.40}} \\
\cmidrule(r){1-2}\cmidrule(lr){3-7}
\multirow{4}{*}{PU} &
VPU~\cite{vpu} & 93.84\err{0.88} & 91.90\err{0.22} & 72.12\err{1.05} & 87.50\err{1.05} & \phantom{00}\na{} \\
& nnPU~\cite{Kiryo2017} &
95.70\err{0.11} &
90.93\err{0.26} &
72.48\err{0.83} &
90.49\err{0.15} &
79.62\err{0.65} 
\\
& \quad +PL &
95.91\err{0.23} &
\textbf{91.36\err{0.13}} &
74.30\err{0.68} &
90.23\err{0.07} &
79.04\err{0.39} \\
& \quad +PUUPL &
\textbf{97.12\err{0.07}} &
91.26\err{0.26} &
\textbf{77.49\err{0.33}} &
\textbf{90.74\err{0.15}} &
\textbf{80.26}\err{0.51} \\
\bottomrule \\
\end{tabular}
}
\caption{\small Average test accuracy and its standard error over five repetitions on various datasets with 3,000 labeled training positives.
The row ``+PL`` refers to an uncertainty-unaware pseudo-labeling baseline, while ``+PUUPL`` refers to our uncertainty-aware solution.
The validation column refers to the use of a fully-labeled (PN) or PU validation set.
}
\label{tbl:test_acc_3k}
\end{table*}

Table~\ref{tbl:imb_acc} shows the performance of PUUPL and the other baselines in an imbalanced scenario with only 600 labeled positives and a prior $\pi=0.1$.
PUUPL was the overall best performer in all comparisons except on the MNIST dataset with PU validation, where its performance was 0.31 percentage points (p.p.) lower than the imbnnPU baseline.
The increase in accuracy was statistically significant with $p<0.0015$ in each case except for CIFAR-10 with PU validation where $p=0.055$.
On CIFAR-10 with PU validation, the improvement against the second-best was not statistically significant, while in all other cases it was with $p<0.0015$.
PUUPL improved performance the most in the IMDb dataset, where accuracy was 2.2 and 3.5 p.p. higher with PN and PU validation sets, and the improvement in CIFAR-100-20 was similarly high with 2.0 and 2.8 p.p. respectively.
Self-PU struggled in this setting, collapsing to negative predictions on CIFAR-100-20 and demonstrating unstable behavior on CIFAR-10, where the collapse only occurred in certain training-validation splits.
The naive pseudo-labeling baseline that did not use uncertainty worsened performance, compared to imbnnPU, in three datasets out of five, regardless of the method used for validation. 
This is, hypothetically, due to the emergence of the confirmation bias.

We performed the same comparison using 3,000 labeled training positives and the natural prior of each dataset, while using the nnPU loss as $\mathcal{L}_{PU}$ (Table~\ref{tbl:test_acc_3k}), as well as 1,000 labeled positives (Table~\ref{tbl:test_acc_1k}).
The results were qualitatively similar, with PUUPL providing the highest test accuracy except for Fashion-MNIST, and sometimes considerable performance increase, for example almost 5 p.p. more in the case of CIFAR-100-20 with 3,000 positives compared to the nnPU baseline.
For Fashion-MNIST, there was no statistically significant difference in accuracy between PUUPL and the uncertainty-unaware pseudo-labeling baseline, while in all other cases the improvement was statistically significant ($p=0.023$ for IMDb with PN validation, $p=0.014$ for CIFAR-10 with PU validation, and $p<0.001$ in all other cases). 

These findings substantiate the advantages of pseudo-labeling in PUL as well as the necessity of uncertainty quantification in this procedure and in particular the benefit that this brings in more imbalanced scenarios with few labeled positives or low prior.

\subsubsection{PUUPL is loss-agnostic}
Our framework is uniquely positioned to take advantage of newly developed risk estimators for PU learning: as we showed above, PUUPL could make use of the imbnnPU loss~\cite{imbnnpu}  and the nnPU loss ~\cite{Kiryo2017} to substantially improve on the state-of-the-art in the imbalanced and the general setting.
Next to imbnnPU, there exists a variety of alternative PU losses for different scenarios.
The nnPUSB loss~\cite{Kato2019} was developed to address the issue of labeling bias in the training positives, a more general setting compared to the i.i.d. assumption of traditional PUL methods~\cite{bekker2020}.
We tested PUUPL in such a biased setting where positives in the CIFAR-10 training and validation sets were with 50\% chance an airplane, 30\% chance an automobile, 15\% chance a ship, and 5\% chance a truck.
The test distribution was instead balanced, meaning for instance that test samples were half as likely to be airplanes compared to the training set, and five times more likely to be truck images.
We used the same hyperparameters as for the i.i.d. CIFAR-10 experiments except for the loss $\mathcal{L}_{PU}$ where we used the nnPUSB loss~\cite{Kato2019} to handle the positive bias.
The baseline with nnPUSB loss performed better than the nnPU loss but worse than \textit{PUUPL} with the nnPU loss, and the best performance was achieved with \textit{PUUPL} on top of the nnPUSB loss (Table~\ref{tbl:bias}).
The improvement provided by PUUPL was statistically significant ($p<1.07\times 10^{-6}$ in all cases).

These results demonstrate that PUUPL can be applied even when a sampling bias is suspected by a practitioner and no \emph{ad hoc} risk estimator is available, as our uncertainty-aware pseudo-labeling framework with the bias-oblivious nnPU loss obtained better results compared to a bias-aware risk estimator without pseudo-labeling.

\begin{table}[t]
\centering
\begin{tabular}{r c c}
\toprule
& \multicolumn{2}{c}{PU loss} \\
&  nnPU~\citep{Kiryo2017} & nnPUSB~\citep{Kato2019} \\
\cmidrule(r){1-1}\cmidrule{2-3}
Only PU loss & 87.05 \err{0.14} & 87.31 \err{0.12}\\
PU loss+PUUPL & \textbf{87.70} \err{0.14} & \textbf{87.91} \err{0.14} \\
\bottomrule \\
\end{tabular}
\caption{\small Test accuracy of PUUPL on the CIFAR-10 dataset with a selection bias on the positive labels when using the nnPU and nnPUSB losses.
Our framework improved over the base PU loss in both cases, and, in particular, PUUPL with nnPU loss achieved better performance than the nnPUSB loss alone.
}
\label{tbl:bias}
\end{table}

\subsubsection{Uncertainty quantification improves pseudo-labels}

\begin{figure}[t]
    \centering
    \includegraphics[width=0.9\linewidth]{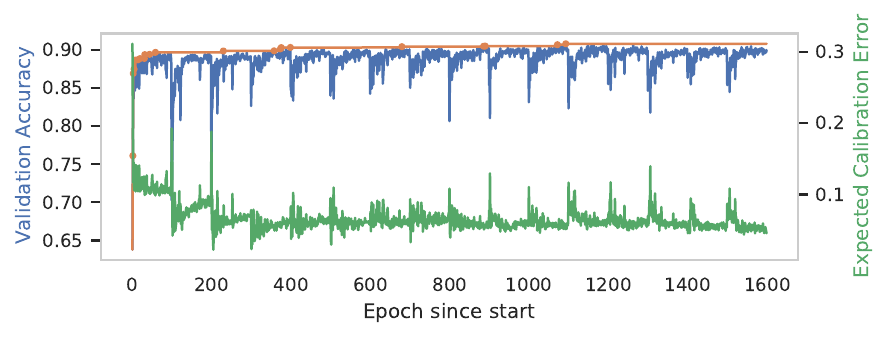}
    \caption{
    \small
    Validation accuracy (left, blue) and expected calibration error (ECE, right, green) for a run on CIFAR-10 with 1,000 positives.
    Note the substantial reduction in ECE in the second and third pseudo-labeling iterations, when the ensemble is trained on soft labels.
    The orange line indicates the best validation accuracy at each epoch, with the new highest accuracy marked by orange dots.
    The overall highest was 90.76\% at epoch 1092, corresponding to a test accuracy of 90.35\%.
    }
    \label{fig:valacccali}
\end{figure}

According to the results in Table~\ref{tbl:imb_acc} and~\ref{tbl:test_acc_3k}, naive pseudo-labeling frequently reduces performance, rather than improving it; it then follows that the performance improvement of PUUPL stems from the uncertainty ranking used to select and assign pseudo-labels (Eq.~\ref{eq:lnew}).
We investigated this in a series of experiments with PUUPL, nnPU, and the naive pseudo-labeling baseline with 1,000 labeled positives, and we found that the improvement in expected calibration error (ECE) on the test set and negative log-likelihood (NLL) of the pseudo-labels assigned by PUUPL was at least 40\% and often much larger (Table~\ref{tbl:ece}).
As shown in Figure~\ref{fig:valacccali}, the ECE decreased during the first few pseudo-labeling rounds, after which it stabilized while the accuracy continued improving.
We also observed that a larger improvement in pseudo-label quality corresponds to a larger improvement in predictive performance.

\begin{table}[t]
\centering
\begin{tabular}{@{} r c c c @{}}
\toprule
&\multicolumn{3}{c}{Test Expected Calibration Error (\%)} \\
& nnPU & +PL & +PUUPL \\
\cmidrule(r){1-1}\cmidrule(l){2-4}
IMDb & 25.94\err{0.78} & 25.23\err{0.11} & \textbf{6.33\err{0.17}} \\
CIFAR-10 & 10.89\err{0.10} & 9.24\err{0.15} & \textbf{5.70\err{0.72}} \\
CIFAR-100-20 & 31.62\err{0.29} & 27.51\err{0.59} & \textbf{22.12\err{0.39}} \\
\cmidrule(r){1-1}\cmidrule(l){2-4}
&\multicolumn{3}{c}{Pseudo-labels Negative Log-Likelihood} \\
& nnPU & +PL & +PUUPL \\
\cmidrule(r){1-1}\cmidrule(l){2-4}
IMDb & \phantom{0.00}\na{} & 2.74\err{0.19} & \textbf{0.61\err{0.02}}  \\
CIFAR-10 & \phantom{0.00}\na{} & 0.66\err{0.05} & \textbf{0.29\err{0.03}} \\
CIFAR-100-20 & \phantom{0.00}\na{} & 3.76\err{0.29} & \textbf{0.94\err{0.05}} \\
\bottomrule \\
\end{tabular}
\caption{\small 
Expected calibration error (ECE) on the test set using 1,000 labeled positives for training (average and standard error over five runs), as well as negative log-likelihood of the assigned pseudo-labels against the true labels.
}
\label{tbl:ece}
\end{table}

\subsubsection{Robustness of PUUPL}
\label{sec:robust}

\paragraph{Prior misspecification}
An important concern for practitioners is how to determine the prior $\pi$ of a PU dataset, as in the case of sub-optimal estimation the performance of the PU classifier can be harmed considerably.
Prior estimation constitutes a whole research branch in PUL \cite{vpu,gargMEP} and is a significant challenge in any practical PU application~\cite{bekker2020}.
Some contemporary methods for PUL~\cite{selfpu,punce,distpu,apu} assume a known prior and do not discuss the practical consequences of not knowing such parameter, while other methods incorporate prior estimation directly into the training procedure~\cite{vpu,gargMEP,Hu2021PredictiveAL}.
We treated $\pi$ as a hyperparameter optimized using the AUROC on a PU validation set as a criterion~\cite{Menon2015LearningFC,Jain2017RecoveringTC}, thus bridging the gap between estimating the prior during training and assuming it is known \emph{a priori}.

Our experimental results show that optimizing the prior in such a way resulted in a consistent reduction in test accuracy between 0.8 and 1.2 percentage points for our framework, the PU loss, and the naive pseudo-labeling baseline (PU sections in Tables~\ref{tbl:imb_acc} and~\ref{tbl:test_acc_3k}).
In a similar vein, methods such as VPU that optimize the prior as part of the training procedure show a similar or larger difference compared to methods that use a PN labeled validation set such as Self-PU.
However, in most cases, PUUPL remained the top performer in both settings.

\begin{figure*}[t]
    \centering
    \includegraphics[width=\linewidth]{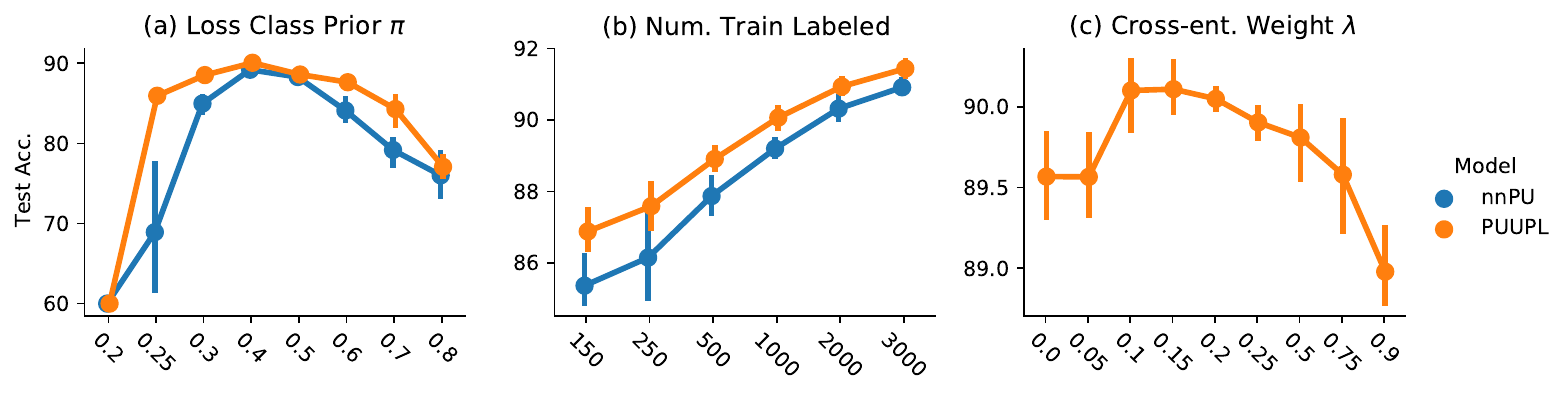}
    \caption{\small Mean and standard deviation of the CIFAR-10 test accuracy obtained over five runs when training with wrong prior (a), number of training labeled positives (b) and
    different loss combination parameter $\lambda$ (c).
    PUUPL proved to be more robust to prior misspecification (true $\pi=0.4$), as the performance degradation was considerably reduced over a wide range of values. 
    It was also more robust to the lower number of labeled samples, as the gap between our framework and nnPU widened when fewer labeled positives were available for training (note the different $y$-axes scales).
    }
    \label{fig:robustness}
\end{figure*}

Moreover, training using a wrong value for $\pi$ was less harmful to PUUPL compared to nnPU only (Fig.~\ref{fig:robustness}a).
For example, on CIFAR-10 the test accuracy showed a wide plateau around the true prior of 0.4 with a performance reduction of less than 2.5\% in the range $[0.3,0.6]$.
With smaller priors, the nnPU loss collapsed to constantly predicting the majority class, and specialized oversampled risk estimators~\cite{imbnnpu} were needed to cope with such a setting (we showed the effectiveness of PUUPL in imbalanced settings in the previous section).
Furthermore, the performance gap between PUUPL and nnPU widened as $\pi$ was more severely misspecified, indicating a higher degree of robustness.

\paragraph{Number of labeled training positives}
The performance of PUUPL steadily increased and seemed to plateau at 91.4\% at 3,000 labeled positives (Fig.~\ref{fig:robustness}b).
The gap between nnPU and PUUPL was largest in the low labeled data region with a 1.44\% gap at 250 labels, where PUUPL achieved 87.59\% accuracy, shrinking to a gap of 0.52\% with 3,000 labels, where PUUPL's performance was 91.44\%.
This supports our intuition about the importance of accounting for prediction uncertainty because, as the amount of labeled data decreases, uncertainty becomes more important to detect overfitting and prevent the model from assigning incorrect pseudo-labels.

\paragraph{Loss mixing parameter}
As a loss, PUUPL uses a convex combination of a loss for the assigned pseudo-labels and the remaining PU data using a mixing coefficient $\lambda$ (Eq.~\ref{eq:combined_loss}).
The best performing combination used $\lambda=0.1$, with modest performance reduction until $\lambda=0.5$ (Fig.~\ref{fig:robustness}c), with too small values nullifying the effect of pseudo-labeling, and larger values harming performance.
In general, when too few samples are pseudo-labeled, the loss $\mathcal{L}_L$ is a high variance estimator of the classification risk, and thus should not be weighted excessively.
This effect may be reduced as more pseudo-labels are added, and dynamic adaptation of $\lambda$ over training could provide an additional performance improvement.

\subsection{Real-world application}

\begin{figure*}[t]
    \centering
    \includegraphics[width=\linewidth]{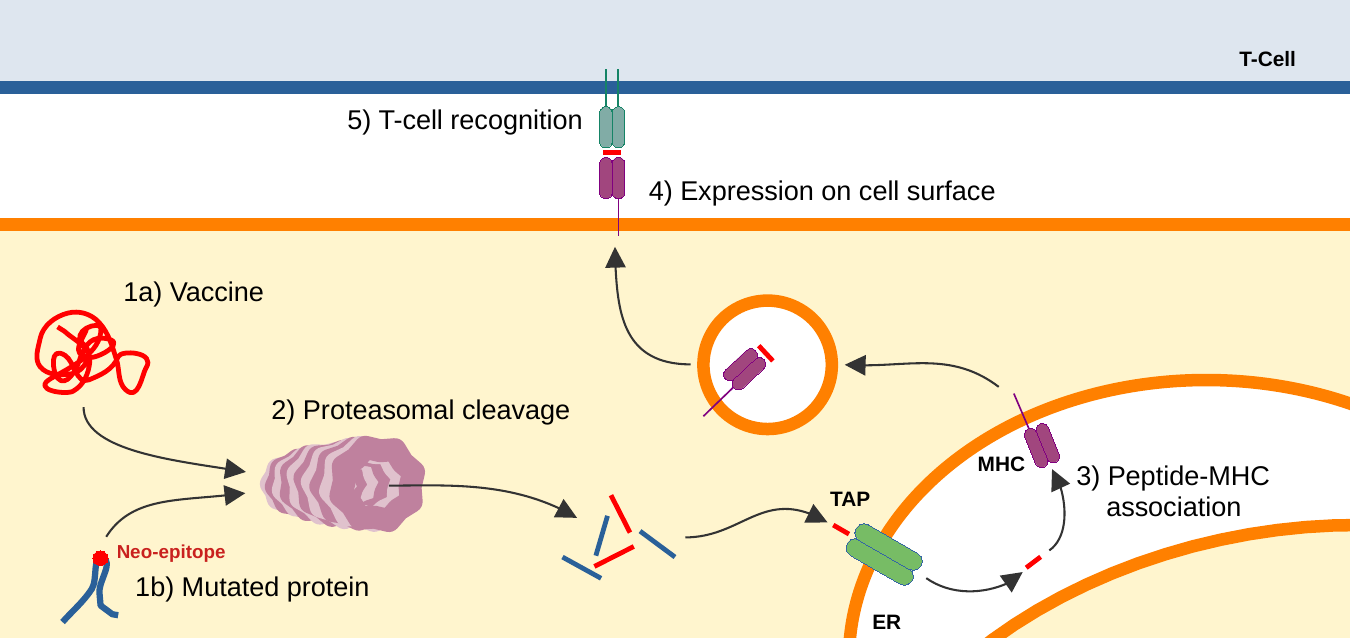}
    \caption{
    \small
    Predicting the outcome of each event in the antigen processing pathway~\citep{antigen_pathway} is crucial to enable the design of epitope vaccines.
    Vaccines ingested by antigen presenting cells (1a) as well as mutated proteins produced by cancerous cells (1b) are cleaved in short fragments by the proteasome (2).
    Some of these fragments, or peptides, are then transported into the endoplasmic reticulum (ER) through the Transporter associated with Antigen Processing (TAP).
    A fraction of these peptides bind to the Major Histocompatibility Complex (MHC, 3) and the resulting construct is then expressed on the cell surface (4), where they can be inspected by passerby T-cells and possibly trigger an appropriate immune response (5).
    }
    \label{fig:apc}
\end{figure*}

In this section we show the applicability and benefits of PUUPL to a real-world imbalanced dataset with applications related to healthcare, improving the predictive performance of previous methods developed \emph{ad hoc}.
Cancer is the result of malignant mutations that were not wiped out in time by the immune system.
It is however possible to instruct the immune system to fight the tumor through specific vaccines that contain neo-epitopes that arose as a result of those mutations, i.e., short genomic regions surrounding the mutated sites that can trigger an immune response~\citep{hu_towards_2017}.
Such vaccines can be designed computationally by solving an optimization problem that chooses the most promising mutations to target while ensuring that the vaccine can be processed appropriately by the body~\cite{jessev,genev,toussaint_universal_2011}.
One of the main steps of such processing~\cite{antigen_pathway} is the digestion of the vaccine by the proteasome, a tubular protein complex that degrades old or misfolded proteins into shorter fragments (Fig.~\ref{fig:apc}).
In order for the vaccine to be effective these pieces must correspond to the neoantigens originally contained in the vaccine. 
Therefore, accurately predicting proteasomal cleavage, i.e., the position where a sequence is cut, is very important to design more effective vaccines.
Modern high-throughput pipelines~\citep{mass_spec} are able to detect MHC-presented epitopes on the cell surface (Fig.~\ref{fig:apc}) which must have originated from proteasomal cleavage.
While missed cleavage sites are never measured, not all presented epitopes are detected, and not all peptides resulting from proteasomal cleavage are presented.
Thus, PUL is a natural abstraction of proteasomal cleavage prediction.

\subsubsection{Dataset}
We collected a dataset of 294,615 MHC-I epitopes from the IEDB~\cite{iedb} database and 89,853 from the Human MHC Ligand Atlas~\cite{hlala}.
To identify the potential progenitor protein of each epitope, we used BLAST~\citep{blast} and filtered for epitopes with a unique progenitor protein resulting in a total of 258,424 data points.
Through the progenitor protein, we recovered the residues preceding the N-terminus and following the C-terminus of the epitope, thus providing context for the cleavage predictor.
We generated two separate datasets based exclusively on N- or C-termini cleavage sites, as it is known that the biological signal differs in these two situations~\citep{Schatz2008}.
We generated ``decoy`` samples by considering cleavage sites located within three residues of the experimentally-determined terminus; as discussed previously, it is unknown whether cleavage could or could not have happened at those positions, hence we treat such decoys as unlabeled in our PUL training procedure.
The final datasets were then composed of 1,285,659 samples with 229,163 positives for the N-terminus datasets and 1,277,344 samples with 222,181 positives for the C-terminus datasets.

\subsubsection{Modeling, Training, and Evaluation}
Each sequence contains ten amino acids, each of which was one-hot encoded and processed by a MLP.
We used imbnnPU~\citep{imbnnpu} as $\mathcal{L}_{PU}$.
For imbnnPU and PUUPL we report the cross-validation scores and use the statistical test proposed by~\citep{LeDell2015ComputationallyEC} to estimate the AUROC, its standard error, and confidence intervals. 
Note that, as we do not know the true negatives, traditional metrics to evaluate classification performance such as accuracy, F1, precision, recall, etc. are not applicable.
As external baselines we consider NetChop~\citep{Nielsen_2005} and NetCleave~\citep{netcleave}, evaluating their predictions on ten random bootstraps of our dataset.
These baselines are based on MLPs and convolutional neural networks respectively and, importantly, they approach the problem as a supervised binary classification task, treating decoy samples as negatives rather than unlabeled.
We also present evaluation scores for the imbnnPU loss~\cite {imbnnpu}, commonly used for PUL on imbalanced datasets.

\begin{table*}[t]
\centering
\begin{tabular}{rrr}
\toprule
&  \multicolumn{2}{c}{AUROC} \\
& \multicolumn{1}{c}{N-terminal} & \multicolumn{1}{c}{C-terminal} \\
\cmidrule(l){1-1}\cmidrule(l){2-3}
NetChop 20S & 52.72\err{0.02} & 66.07\err{0.02} \\
NetChop C term & 50.99\err{0.02} & 81.53\err{0.01} \\
NetCleave & 49.27\err{0.02} &  79.61\err{0.01} \\
\cmidrule(l){1-1}\cmidrule(l){2-3}
imbnnPU & 75.15\err{0.06} & 83.99\err{0.06} \\
PUUPL & \textbf{78.00}\err{0.06} & \textbf{87.20}\err{0.04} \\
\bottomrule \\
\end{tabular}
\caption{\small 
Average and standard error of area under the ROC curve (AUROC) on both datasets for NetChop, NetCleave, the imbnnPU loss and PUUPL.
}
\label{tbl:proc}
\end{table*}

\subsubsection{Results}
Both PUUPL and the imbnnPU loss achieved lower performance on the N-terminals dataset, confirming previous observations that this predictive task is harder due to the biological processes involved~\citep{Schatz2008}.
On the C-terminal dataset, the imbnnPU loss improved performance by 2.5 and 4.4 points compared to NetChop and NetCleave respectively, and PUUPL added a further 3.2 points reaching 87.2\% AUROC (Table~\ref{tbl:proc}).
In both datasets the difference in AUROC between imbnnPU and PUUPL was statistically significant at a significance of 1\%: the confidence intervals are $[74.99,75.32]$ and $[77.85,78.15]$ for N-terminals, and  $[83.85,84.14]$ and $[87.08,87.32]$ for C-terminals.
Note that both NetChop and NetCleave were only trained on C-terminals cleavage sites in the original publication, thus explaining their random predictions on the N-terminals dataset.

\section{Discussion and conclusions}
We introduced PUUPL, an uncertainty-aware pseudo-labeling framework for PUL that uses the epistemic uncertainty of an ensemble of networks to select which examples to pseudo-label.
We conducted extensive experiments to demonstrate the benefits of our approach and show its reliability in settings that are likely to be encountered in the real world such as heavily imbalanced settings with small $\pi$ and few labeled positives, a bias in the positive training data, the unavailability of labeled negatives for validation, and the misspecification of the class prior $\pi$.
Unlike many alternative methods, PUUPL can be applied to learning problems in any domain out of the box as it does not rely on regularization methods that are restricted to a specific data modality, most frequently images, such as mixup~\cite{mixup} (used by~\cite{vpu,distpu}) or contrastive representations (used by~\cite{punce}). 
Furthermore, it is easy to adapt as it builds on standard methods (unlike~\cite{selfpu}), and does not require pretrained representations to work (as~\cite{apu} does).
We further used our framework to advance the state-of-the-art on a real-world healthcare dataset with potential repercussions on efficacy and deployment cost of personalized epitope vaccines for cancer treatment.

Our choice of deep ensembles was rooted in their competitiveness in empirical benchmarks~\cite{ovadia2019can}. 
However, PUUPL can easily be extended to take advantage of more accurate uncertainty quantification methods as they become available~\cite{abdar2021review}, especially considering the computational overhead of training deep ensembles.
In fact, as the matter of uncertainty quantification in deep learning is far from settled, the performance and efficiency of our framework could be further improved by employing more accurate uncertainty quantification methods~\cite{abdar2021review}.
We demonstrated robustness against biased positive labels and imbalanced datasets, however, it is the practitioners' responsibility to ensure that the obtained predictions are ''fair'', with ''fairness'' defined appropriately with respect to the target application, and do not systematically affect particular subsets of the population of interest.
Ultimately, we can only leave it to practitioners to use their moral and ethical judgment as to whether all stakeholders and their interests are fairly represented in their application.

To conclude, our PUUPL framework is applicable to and reaches competitive performance in heavily imbalanced settings with few positives as well as in traditional balanced datasets, even when the true prior is unknown, and can work with several data modalities including images, natural language, and proteomic data while being robust to hyperparameter selection.


\subsection*{Declarations}
\noindent\textbf{Acknowledgments}
We are grateful to David R\"ugamer, Niki Kilbertus, Bastian Rieck and the anonymous reviewers for their comments and suggestions.

\noindent\textbf{Funding:}
Emilio Dorigatti was supported by the Helmholtz Association under the joint research school "Munich School for Data Science - MUDS" (Award Number HIDSS-0006).
Jann Goschenhofer was supported by the Bavarian Ministry of Economic Affairs, Regional Development, and Energy through the Center for Analytics – Data – Applications (ADA-Center) within the framework of BAYERN DIGITAL II (20-3410-2-9-8).
B. S. acknowledges financial support by the Postdoctoral Fellowship Program of the Helmholtz Zentrum M\"unchen. Mina Rezaei and Bernd Bischl were supported by the German Federal Ministry of Education and Research (BMBF) under Grant No. 01IS18036A, Munich Center for Machine Learning (MCML). 

\noindent\textbf{Conflicts of interest/Competing interests:}
The authors of the manuscript do not have a conflict of interest. 

\noindent\textbf{Authors' contributions:}
The method was conceived by E.D. and finalized upon discussion with all other authors.
All authors contributed to the experimental protocol, while the implementation was performed by E.D. and J.G., who also performed the experiments.
All authors contributed to the interpretation of the results.
The manuscript was written by E.D. and J.G. with feedback from all other authors.
All authors read and approved the final manuscript.

\newpage

\begin{appendices}

\section{Network architecture, hyperparameters and datasets}
\label{sec:hypars}
Table~\ref{tbl:dsets} reports the number of samples in each dataset.
Table~\ref{tbl:c10net} reports the network architecture used in the CIFAR-10 and CIFAR-100-20 experiments, while Table~\ref{tbl:imdbnet} reports the network used with IMDb.
Table~\ref{tbl:hpars} reports the hyperparameters related to pseudo-labeling and their ranges.

\begin{table}
\centering
\begin{tabular}{rccc}
\toprule
Dataset & Train Pos. & Train Neg. & Test Size \\
\cmidrule(r){1-1}\cmidrule{2-4}
MNIST & 30,508 & 29,492 & 10,000 \\
F-MNIST & 30,000 & 30,000 & 10,000 \\
CIFAR-10 & 20,000 & 30,000 & 10,000 \\
CIFAR-100-20 & 25,000 & 25,000 & 10,000 \\
IMDb & 12,500 & 12,500 & 25,000 \\
\bottomrule \\
\end{tabular}
\caption{Size of test set and number of positives and negatives in the training set for each dataset.}
\label{tbl:dsets}
\end{table}

\begin{table*}
\centering
\begin{tabular}{ r | l}
\textbf{Layer type} & \textbf{Layer parameters} \\
\hline
Conv. 2D & InC=3, OutC=96, k=3, s=1, p=1 \\
Dropout & p=0.15 \\
Batch Norm. & eps=1e-05, momentum=0.1\\
ReLU &  \\
\hline
Conv. 2D & InC=96, OutC=96, k=3, s=1, p=1 \\
Dropout & p=0.15\\
Batch Norm. & eps=1e-05, momentum=0.1\\
ReLU &  \\
\hline
Conv. 2D & InC=96, OutC=96, k=3, s=2, p=1 \\
Dropout & p=0.15\\
Batch Norm. & eps=1e-05, momentum=0.1\\
ReLU &  \\
\hline
Conv. 2D & InC=96, OutC=192, k=3, s=1, p=1 \\
Dropout & p=0.15\\
Batch Norm. & eps=1e-05, momentum=0.1\\
ReLU &  \\
\hline
Conv. 2D & InC=192, OutC=192, k=3, s=1, p=1 \\
Dropout & p=0.15\\
Batch Norm. & eps=1e-05, momentum=0.1\\
ReLU &  \\
\hline
Conv. 2D & InC=192, OutC=192, k=3, s=2, p=1 \\
Dropout & p=0.15\\
Batch Norm. & eps=1e-05, momentum=0.1\\
ReLU &  \\
\hline
Conv. 2D & InC=192, OutC=192, k=3, s=1, p=1 \\
Dropout & p=0.15\\
Batch Norm. & eps=1e-05, momentum=0.1\\
ReLU &  \\
\hline
Conv. 2D & InC=192, OutC=192, k=1, s=1, p=0 \\
Dropout & p=0.15\\
Batch Norm. & eps=1e-05, momentum=0.1\\
ReLU &  \\
\hline
Conv. 2D & InC=192, OutC=10, k=1, s=1, p=0 \\
Dropout & p=0.15\\
Batch Norm. & eps=1e-05, momentum=0.1\\
ReLU &  \\
\hline
Flatten & \\
\hline
Linear & in features=640, out features=1000, bias=yes \\
ReLU &  \\
\hline
Linear & in features=1000, out features=1000, bias=yes \\
ReLU &  \\
\hline
Linear & in features=1000, out features=1, bias=yes \\
\end{tabular}
\caption{
Network architecture used for the CIFAR-10 experiments.
InC: input channels, OutC: output channels, k: kernel size, s: stride, p: padding
}
\label{tbl:c10net}
\end{table*}

\begin{table}
\centering
\begin{tabular}{ r | l}
\textbf{Layer type} & \textbf{Layer parameters} \\
\hline
LSTM & inpus size=200, hidden size=128, \\
& num layers=2, dropout=0.25, \\
& bidirectional=True \\
Dropout & p=0.2\\
\hline
Linear & in features=256, out features=196, \\
& bias=True \\
Batch Norm. & eps=1e-05, momentum=0.1 \\
ReLU & \\
Dropout & p=0.2\\
\hline
Linear&in features=196, out features=196, \\
& bias=True \\
Batch Norm.&eps=1e-05, momentum=0.1\\
ReLU& \\
Linear&in features=196, out features=1, \\
& bias=True \\
\end{tabular}
\caption{Network architecture used for the IMDb experiments}
\label{tbl:imdbnet}
\end{table}

\begin{table}
\centering
\begin{tabular}{ r | l}
\textbf{Hyper-parameter} & \textbf{Value range} \\
\hline
Estimator & Ensemble or MC Dropout \\
Number of samples & $[2, 25]$ \\
Uncertainty type & Aleatoric, epistemic, total \\
Max. new labels $T$ & $[100, 5000]$ \\
Max. new label uncertainty $t_l$ & $[0, -\log 2]$ \\
Min. unlabel uncertainty $t_u$ & $[0, -\log 2]$ \\
Reassign all pseudo-labels & Yes or no \\
Re-initialize to same weights & Yes or no \\
Cross-entropy weight $\lambda$ & $[0, 1]$ \\
\end{tabular}
\caption{Pseudo-labeling hyperparameters}
\label{tbl:hpars}
\end{table}

\section{Further results and sensitivity analyses}
\label{sec:apx:sensitivity}

\begin{table*}
\resizebox{\textwidth}{!}{%
\centering
\begin{tabular}{ l l l l l l l }
\toprule
& &
\multicolumn{5}{c}{Dataset } \\
\cmidrule{3-7}
Valid. &
Method &
MNIST &
F-MN  &
C-100-20  &
CIF-10  & 
IMDb \\
\cmidrule(r){1-2}\cmidrule(lr){3-7}
\multirow{4}{*}{PN} &
Self-PU~\cite{selfpu} & 93.09\err{0.16} & 90.21\err{0.15} & 70.56\err{0.62} & 88.33\err{0.18} & \phantom{00}\na{} \\
& nnPU~\cite{Kiryo2017} &
94.68\err{0.13} &
90.55\err{0.08} &
67.69\err{0.23} &
88.87\err{0.17} &
71.45\err{0.38}
\\
& \quad +PL             &
95.32\err{0.24} &
90.69\err{0.09} &
69.01\err{0.56} &
\textbf{90.02\err{0.09}} &
71.70\err{0.14} \\
& \quad +PUUPL &
\textbf{96.09\err{0.08}} &
\textbf{90.92\err{0.07}} &
\textbf{71.44\err{0.37}} &
89.84\err{0.13} &
\textbf{74.14\err{0.35}} \\
\cmidrule(r){1-2}\cmidrule(lr){3-7}
\multirow{4}{*}{PU} &
VPU~\cite{vpu} & 89.78\err{0.76} & 85.50\err{0.63} & 69.58\err{1.47} & 85.06\err{0.55} & \phantom{00}\na{} \\
& nnPU~\cite{Kiryo2017} &
93.89\err{0.38} &
89.53\err{0.61} &
67.25\err{0.98} &
87.73\err{0.32} &
71.43\err{0.39} 
\\
& \quad +PL &
92.16\err{0.50} &
\textbf{90.63\err{0.33}} &
67.93\err{0.49} &
\textbf{88.92\err{0.19}} &
71.44\err{0.28} \\
& \quad +PUUPL &
\textbf{94.85\err{0.08}} &
90.05\err{0.17} &
\textbf{69.92\err{0.70}} &
88.82\err{0.16} &
\textbf{73.98\err{0.25}} \\
\bottomrule \\
\end{tabular}
}
\caption{\small Average test accuracy and its standard error over five repetitions on various datasets with 1,000 labeled training positives.
The row ``+PL`` refers to an uncertainty-unaware pseudo-labeling baseline, while ``+PUUPL`` refers to our uncertainty-aware solution.
The validation column refers to the use of a fully-labeled (PN) or PU validation set.
}
\label{tbl:test_acc_1k}
\end{table*}

Table~\ref{tbl:test_acc_1k} reports the test accuracy when 1,000 labeled positives were used for training.

We performed ablation studies on the CIFAR-10 dataset by changing one parameter at a time of the best configuration found by Hyperband, training and evaluating with five different splits, and reporting the test accuracy corresponding to the best validation score for each run.
To limit the computational resources needed, we used at most 15 pseudo-labeling iterations.

\begin{figure*}
    \centering
    \includegraphics[width=\linewidth]{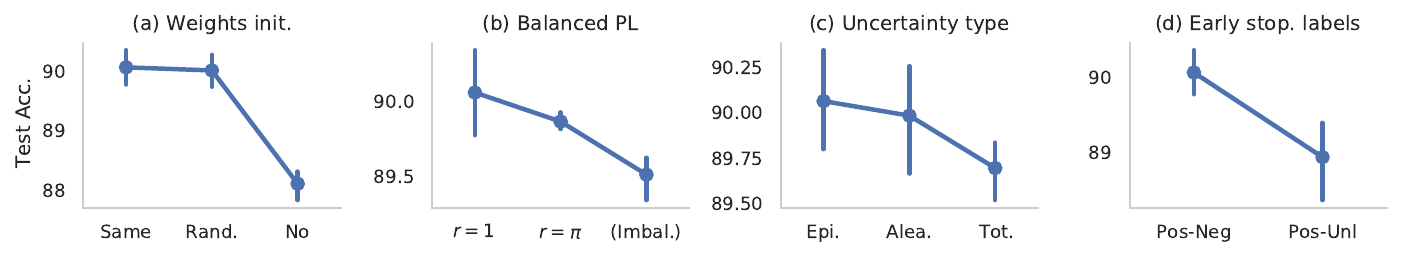}
    \caption{
    \small
    Mean and standard deviation of the test accuracy obtained over five runs by different variations of our \textit{PUUPL} algorithm: (a) different weight initialization at each iteration, (b) balanced or imbalanced PL selection, (c) type of uncertainty, (d) whether to use PN or PU validation set. Note the different scales on the $y$-axes.}
    \label{fig:algos}
\end{figure*}

\textbf{Weight initialization:}
We confirmed the observation that it is beneficial to re-initialize the weights after each pseudo-labeling step \cite{Arazo2020PseudoLabelingAC}, with slightly better performance ($+0.052\%$) achieved when the weights are re-initialized to the same values before every pseudo-labeling iteration (Fig.~\ref{fig:algos}a).
We believe this encourages the model to be consistent across pseudo-labeling rounds.

\textbf{Uncertainty:}
Ranking predictions by aleatoric performance was almost as good as ranking by epistemic uncertainty ($-0.08\%$), while total uncertainty produced moderately worse rankings ($-0.37\%$, Fig.~\ref{fig:algos}c).
An ensemble with only two networks achieved the best performance, while larger ensembles performed worse, and Monte Carlo dropout ($-0.85\%$) was better than ensembles of five ($-1.00\%$) and ten networks ($-1.58\%$).

\textbf{Early stopping:}
Finally, performing early stopping on the validation PU loss resulted in worse accuracy ($-1.12\%$) compared to using the accuracy on PN labels (Fig.~\ref{fig:algos}d).
Although considerable when compared to the impact of other algorithmic choices, such a performance drop indicates that \textit{PUUPL} can be used effectively in real-world scenarios when no labeled validation data are available.

\textbf{Pseudo-labeling hyperparameters:}
Our method was fairly robust to the maximum number $T$ of assigned pseudo-labels and the maximum uncertainty threshold $t_l$ for the pseudo-labels, with almost constant performance up to $T=1000$ and $t_l=0.1$.
The best performance was achieved by the combination having $T=1000$ and $t_l=0.05$, but both of these experiments were performed while disabling the other constraint (i.e., setting $T=\inf$ when testing $t_l$ and vice-versa).
Using only a constraint on $T$ resulted in a reduction of $-0.11\%$, while constraining $t_l$ alone resulted in a reduction of $-1.04\%$.
The results for $t_u$ were less conclusive than for the general trend, possibly because values lower than 0.35 require more than the 15 pseudo-labeling iterations we used for the experiment, and values above 0.4 did not show significant differences.

Moreover, soft pseudo-labels were preferred over hard ones ($+0.75\%$).
Contrary to expectation, however, re-assigning all pseudo-labels at every iteration slighly harmed performance ($-0.12\%$); instead, pseudo-labels should be kept fixed after being assigned for the first time.
A possible explanation is that fixed pseudo-labels prevent the model's predictions from drifting too far away from the initial pseudo-labeling towards an incorrect assignment, and thus contribute in mitigating the sort of confirmation bias that frequently plagues pseudo-labeling-based methods.
It was also beneficial to assign the same number of positive and negative pseudo-labels compared to keeping the same ratio $\pi$ of positives and negatives found in the whole dataset ($-0.20\%$) or not balancing the selection at all ($-0.55\%$).
This prevents the pseudo-labeled set from becoming too imbalanced over time, a natural tendency deriving from the inherent imbalance between positive and unlabeled samples in the training set.

\section{Ethics statement and broader impact}
Improving performance of PUL methods will catalyze research in areas where PU datasets are endemic and manual annotation is expensive or negative samples are impossible to obtain -- for example, in bioinformatics and medical applications -- which ultimately benefits human welfare and well-being.
The explicit incorporation of uncertainty quantification further increases the trustworthiness and reliability of PUUPL's predictions.
However, such advances in PUL could also reduce the resources required to create unwanted mass-surveillance systems by governments and/or private companies.

\end{appendices}

{
\small
\bibliographystyle{plainnat}
\bibliography{main}

\begin{thebibliography}{58}
\providecommand{\natexlab}[1]{#1}
\providecommand{\url}[1]{\texttt{#1}}
\expandafter\ifx\csname urlstyle\endcsname\relax
  \providecommand{\doi}[1]{doi: #1}\else
  \providecommand{\doi}{doi: \begingroup \urlstyle{rm}\Url}\fi

\bibitem[Abdar et~al.(2021)Abdar, Pourpanah, Hussain, Rezazadegan, Liu,
  Ghavamzadeh, Fieguth, Cao, Khosravi, Acharya, et~al.]{abdar2021review}
Moloud Abdar, Farhad Pourpanah, Sadiq Hussain, Dana Rezazadegan, Li~Liu,
  Mohammad Ghavamzadeh, Paul Fieguth, Xiaochun Cao, Abbas Khosravi, U~Rajendra
  Acharya, et~al.
\newblock A review of uncertainty quantification in deep learning: Techniques,
  applications and challenges.
\newblock \emph{Information Fusion}, 76:\penalty0 243--297, 2021.

\bibitem[Acharya et~al.(2022)Acharya, Sanghavi, Jing, Bhushanam, Choudhary,
  Rabbat, and Dhillon]{punce}
Anish Acharya, Sujay Sanghavi, Li~Jing, Bhargav Bhushanam, Dhruv Choudhary,
  Michael~G. Rabbat, and Inderjit~S. Dhillon.
\newblock Positive unlabeled contrastive learning.
\newblock \emph{ArXiv}, abs/2206.01206, 2022.

\bibitem[Altschul et~al.(1990)Altschul, Gish, Miller, Myers, and Lipman]{blast}
Stephen~F. Altschul, Warren Gish, Webb Miller, Eugene~W. Myers, and David~J.
  Lipman.
\newblock Basic local alignment search tool.
\newblock \emph{Journal of Molecular Biology}, 215\penalty0 (3):\penalty0
  403--410, October 1990.
\newblock \doi{10.1016/s0022-2836(05)80360-2}.

\bibitem[Amengual-Rigo and Guallar(2021)]{netcleave}
Pep Amengual-Rigo and Victor Guallar.
\newblock {NetCleave}: an open-source algorithm for predicting c-terminal
  antigen processing for {MHC}-i and {MHC}-{II}.
\newblock \emph{Scientific Reports}, 11\penalty0 (1), June 2021.
\newblock \doi{10.1038/s41598-021-92632-y}.

\bibitem[Arazo et~al.(2020)Arazo, Ortego, Albert, O'Connor, and
  McGuinness]{Arazo2020PseudoLabelingAC}
Eric Arazo, Diego Ortego, P.~Albert, N.~O'Connor, and Kevin McGuinness.
\newblock Pseudo-labeling and confirmation bias in deep semi-supervised
  learning.
\newblock \emph{2020 International Joint Conference on Neural Networks
  (IJCNN)}, pages 1--8, 2020.

\bibitem[Armenian and Lilienfeld(1974)]{armenian1974distribution}
Harotune~K Armenian and Abraham~M Lilienfeld.
\newblock The distribution of incubation periods of neoplastic diseases.
\newblock \emph{American journal of epidemiology}, 99\penalty0 (2):\penalty0
  92--100, 1974.

\bibitem[Bekker and Davis(2020)]{bekker2020}
Jessa Bekker and Jesse Davis.
\newblock Learning from positive and unlabeled data: A survey.
\newblock \emph{Machine Learning}, 109\penalty0 (4):\penalty0 719--760, 2020.

\bibitem[Beluch et~al.(2018)Beluch, Genewein, N{\"u}rnberger, and
  K{\"o}hler]{Beluch2018}
William~H Beluch, Tim Genewein, Andreas N{\"u}rnberger, and Jan~M K{\"o}hler.
\newblock The power of ensembles for active learning in image classification.
\newblock In \emph{Proceedings of the IEEE/CVF Conference on Computer Vision
  and Pattern Recognition}, pages 9368--9377, 2018.

\bibitem[Blum et~al.(2013)Blum, Wearsch, and Cresswell]{antigen_pathway}
Janice~S. Blum, Pamela~A. Wearsch, and Peter Cresswell.
\newblock Pathways of antigen processing.
\newblock \emph{Annual Review of Immunology}, 31\penalty0 (1):\penalty0
  443--473, March 2013.
\newblock \doi{10.1146/annurev-immunol-032712-095910}.

\bibitem[Chapelle et~al.(2009)Chapelle, Scholkopf, and Zien]{Chapelle2009}
Olivier Chapelle, Bernhard Scholkopf, and Alexander Zien.
\newblock Semi-supervised learning.
\newblock \emph{Cambridge, Massachusettes: The MIT Press View Article}, 2009.

\bibitem[Chen et~al.(2020{\natexlab{a}})Chen, Liu, Wang, Zhao, and Wu]{vpu}
Hui Chen, Fangqing Liu, Yin Wang, Liyue Zhao, and Hao Wu.
\newblock A variational approach for learning from positive and unlabeled data.
\newblock In \emph{Advances in Neural Information Processing Systems}, pages
  14844--14854, 2020{\natexlab{a}}.

\bibitem[Chen et~al.(2020{\natexlab{b}})Chen, Chen, Chen, Yuan, Gong, Chen, and
  Wang]{selfpu}
Xuxi Chen, Wuyang Chen, Tianlong Chen, Ye~Yuan, Chen Gong, Kewei Chen, and
  Zhangyang Wang.
\newblock Self-pu: Self boosted and calibrated positive-unlabeled training.
\newblock In \emph{International Conference on Machine Learning}, pages
  1510--1519. PMLR, 2020{\natexlab{b}}.

\bibitem[Christoffel et~al.(2016)Christoffel, Niu, and Sugiyama]{marthinus2016}
Marthinus Christoffel, Gang Niu, and Masashi Sugiyama.
\newblock Class-prior estimation for learning from positive and unlabeled data.
\newblock In \emph{Asian Conference on Machine Learning}, volume~45 of
  \emph{Proceedings of Machine Learning Research}, pages 221--236, 2016.

\bibitem[Deng(2012)]{deng2012mnist}
Li~Deng.
\newblock The mnist database of handwritten digit images for machine learning
  research.
\newblock \emph{IEEE Signal Processing Magazine}, 29\penalty0 (6):\penalty0
  141--142, 2012.

\bibitem[Dorigatti and Schubert(2020{\natexlab{a}})]{genev}
Emilio Dorigatti and Benjamin Schubert.
\newblock Graph-theoretical formulation of the generalized epitope-based
  vaccine design problem.
\newblock \emph{{PLOS} Computational Biology}, 16\penalty0 (10):\penalty0
  e1008237, October 2020{\natexlab{a}}.
\newblock \doi{10.1371/journal.pcbi.1008237}.

\bibitem[Dorigatti and Schubert(2020{\natexlab{b}})]{jessev}
Emilio Dorigatti and Benjamin Schubert.
\newblock Joint epitope selection and spacer design for string-of-beads
  vaccines.
\newblock \emph{Bioinformatics}, 36\penalty0 (Supplement{\_}2):\penalty0
  i643--i650, December 2020{\natexlab{b}}.
\newblock \doi{10.1093/bioinformatics/btaa790}.

\bibitem[Du~Plessis et~al.(2014)Du~Plessis, Niu, and Sugiyama]{plessis2014}
Marthinus~C Du~Plessis, Gang Niu, and Masashi Sugiyama.
\newblock Analysis of learning from positive and unlabeled data.
\newblock \emph{Advances in neural information processing systems},
  27:\penalty0 703--711, 2014.

\bibitem[Garg et~al.(2021)Garg, Wu, Smola, Balakrishnan, and Lipton]{gargMEP}
Saurabh Garg, Yifan Wu, Alexander~J Smola, Sivaraman Balakrishnan, and Zachary
  Lipton.
\newblock Mixture proportion estimation and pu learning:a modern approach.
\newblock In M.~Ranzato, A.~Beygelzimer, Y.~Dauphin, P.S. Liang, and J.~Wortman
  Vaughan, editors, \emph{Advances in Neural Information Processing Systems},
  volume~34, pages 8532--8544, 2021.

\bibitem[Gawlikowski et~al.(2021)Gawlikowski, Tassi, Ali, Lee, Humt, Feng,
  Kruspe, Triebel, Jung, Roscher, Shahzad, Yang, Bamler, and
  Zhu]{Gawlikowski2021ASO}
Jakob Gawlikowski, Cedrique Rovile~Njieutcheu Tassi, Mohsin Ali, Jongseo Lee,
  Matthias Humt, Jianxiang Feng, Anna Kruspe, R.~Triebel, P.~Jung, R.~Roscher,
  M.~Shahzad, Wen Yang, R.~Bamler, and Xiaoxiang Zhu.
\newblock A survey of uncertainty in deep neural networks.
\newblock \emph{ArXiv}, abs/2107.03342, 2021.

\bibitem[Gligorijevi{\'c} et~al.(2021)Gligorijevi{\'c}, Renfrew, Kosciolek,
  Leman, Berenberg, Vatanen, Chandler, Taylor, Fisk, Vlamakis,
  et~al.]{gligorijevic2021structure}
Vladimir Gligorijevi{\'c}, P~Douglas Renfrew, Tomasz Kosciolek, Julia~Koehler
  Leman, Daniel Berenberg, Tommi Vatanen, Chris Chandler, Bryn~C Taylor, Ian~M
  Fisk, Hera Vlamakis, et~al.
\newblock Structure-based protein function prediction using graph convolutional
  networks.
\newblock \emph{Nature communications}, 12\penalty0 (1):\penalty0 1--14, 2021.

\bibitem[Hammoudeh and Lowd(2020)]{apu}
Zayd Hammoudeh and Daniel Lowd.
\newblock Learning from positive and unlabeled data with arbitrary positive
  shift.
\newblock In H.~Larochelle, M.~Ranzato, R.~Hadsell, M.F. Balcan, and H.~Lin,
  editors, \emph{Advances in Neural Information Processing Systems}, volume~33,
  pages 13088--13099, 2020.

\bibitem[Hsieh et~al.(2019)Hsieh, Niu, and Sugiyama]{Hsieh2019}
Yu-Guan Hsieh, Gang Niu, and Masashi Sugiyama.
\newblock Classification from positive, unlabeled and biased negative data.
\newblock In Kamalika Chaudhuri and Ruslan Salakhutdinov, editors,
  \emph{Proceedings of the 36th International Conference on Machine Learning},
  volume~97 of \emph{Proceedings of Machine Learning Research}, pages
  2820--2829, 09--15 Jun 2019.

\bibitem[Hu et~al.(2021)Hu, Le, Liu, Ji, Ma, Zhao, and Yan]{Hu2021PredictiveAL}
Wenpeng Hu, Ran Le, Bing Liu, Feng Ji, Jinwen Ma, Dongyan Zhao, and Rui Yan.
\newblock Predictive adversarial learning from positive and unlabeled data.
\newblock \emph{Proceedings of the AAAI Conference on Artificial Intelligence},
  35\penalty0 (9):\penalty0 7806--7814, May 2021.

\bibitem[Hu et~al.(2017)Hu, Ott, and Wu]{hu_towards_2017}
Zhuting Hu, Patrick~A. Ott, and Catherine~J. Wu.
\newblock Towards personalized, tumour-specific, therapeutic vaccines for
  cancer.
\newblock \emph{Nature Reviews Immunology}, 18\penalty0 (3):\penalty0 168--182,
  2017.
\newblock ISSN 1474-1733, 1474-1741.
\newblock \doi{10.1038/nri.2017.131}.

\bibitem[H{\"u}llermeier and Waegeman(2021)]{Hllermeier2021AleatoricAE}
Eyke H{\"u}llermeier and Willem Waegeman.
\newblock Aleatoric and epistemic uncertainty in machine learning: An
  introduction to concepts and methods.
\newblock \emph{Machine Learning}, 110\penalty0 (3):\penalty0 457--506, 2021.

\bibitem[Iscen et~al.(2019)Iscen, Tolias, Avrithis, and Chum]{Iscen2019}
Ahmet Iscen, Giorgos Tolias, Yannis Avrithis, and Ondrej Chum.
\newblock Label propagation for deep semi-supervised learning.
\newblock In \emph{Proceedings of the IEEE/CVF Conference on Computer Vision
  and Pattern Recognition}, pages 5070--5079, 2019.

\bibitem[Jain et~al.(2017)Jain, White, and Radivojac]{Jain2017RecoveringTC}
Shantanu Jain, Martha White, and Predrag Radivojac.
\newblock Recovering true classifier performance in positive-unlabeled
  learning.
\newblock \emph{Proceedings of the AAAI Conference on Artificial Intelligence},
  31\penalty0 (1), Feb. 2017.

\bibitem[Kaji et~al.(2018)Kaji, Yamaguchi, and Sugiyama]{KajiYS18}
Hirotaka Kaji, Hayato Yamaguchi, and Masashi Sugiyama.
\newblock Multi task learning with positive and unlabeled data and its
  application to mental state prediction.
\newblock In \emph{2018 IEEE International Conference on Acoustics, Speech and
  Signal Processing (ICASSP)}, pages 2301--2305, 2018.
\newblock \doi{10.1109/ICASSP.2018.8462108}.

\bibitem[Kato et~al.(2019)Kato, Teshima, and Honda]{Kato2019}
Masahiro Kato, Takeshi Teshima, and Junya Honda.
\newblock Learning from positive and unlabeled data with a selection bias.
\newblock In \emph{International Conference on Learning Representations}, 2019.

\bibitem[Kingma and Ba(2015)]{Kingma2015AdamAM}
Diederik~P. Kingma and Jimmy Ba.
\newblock Adam: A method for stochastic optimization.
\newblock \emph{CoRR}, abs/1412.6980, 2015.

\bibitem[Kiryo et~al.(2017)Kiryo, Niu, du~Plessis, and Sugiyama]{Kiryo2017}
Ryuichi Kiryo, Gang Niu, Marthinus~C du~Plessis, and Masashi Sugiyama.
\newblock Positive-unlabeled learning with non-negative risk estimator.
\newblock \emph{Advances in Neural Information Processing Systems}, 2017.

\bibitem[Krizhevsky et~al.(2009)Krizhevsky, Hinton,
  et~al.]{krizhevsky2009learning}
Alex Krizhevsky, Geoffrey Hinton, et~al.
\newblock Learning multiple layers of features from tiny images.
\newblock \emph{Citeseer}, 2009.

\bibitem[Lakshminarayanan et~al.(2017)Lakshminarayanan, Pritzel, and
  Blundell]{deepens}
Balaji Lakshminarayanan, Alexander Pritzel, and Charles Blundell.
\newblock Simple and scalable predictive uncertainty estimation using deep
  ensembles.
\newblock In I.~Guyon, U.~V. Luxburg, S.~Bengio, H.~Wallach, R.~Fergus,
  S.~Vishwanathan, and R.~Garnett, editors, \emph{Advances in Neural
  Information Processing Systems}, volume~30, 2017.

\bibitem[LeDell et~al.(2015)LeDell, Petersen, and van~der
  Laan]{LeDell2015ComputationallyEC}
Erin LeDell, Maya~L. Petersen, and Mark~J. van~der Laan.
\newblock Computationally efficient confidence intervals for cross-validated
  area under the roc curve estimates.
\newblock \emph{Electronic journal of statistics}, 9 1:\penalty0 1583--1607,
  2015.

\bibitem[Lee(2013)]{Lee2013}
Dong-Hyun Lee.
\newblock Pseudo-label: The simple and efficient semi-supervised learning
  method for deep neural networks.
\newblock In \emph{Workshop on Challenges in Representation Learning, ICML},
  volume~3, page 896, 2013.

\bibitem[Li et~al.(2010)Li, Guo, and Elkan]{li2010positive}
Wenkai Li, Qinghua Guo, and Charles Elkan.
\newblock A positive and unlabeled learning algorithm for one-class
  classification of remote-sensing data.
\newblock \emph{IEEE Transactions on geoscience and remote sensing},
  49\penalty0 (2):\penalty0 717--725, 2010.

\bibitem[Liu et~al.(2003)Liu, Dai, Li, Lee, and Yu]{Liu2003}
Bing Liu, Yang Dai, Xiaoli Li, Wee~Sun Lee, and Philip~S Yu.
\newblock Building text classifiers using positive and unlabeled examples.
\newblock In \emph{Third IEEE International Conference on Data Mining}, pages
  179--186. IEEE, 2003.

\bibitem[Luo et~al.(2021)Luo, Zhao, Chen, Qiao, Du, Zhang, Wu, Cai, He,
  Rajmohan, et~al.]{Luo2021}
Chuan Luo, Pu~Zhao, Chen Chen, Bo~Qiao, Chao Du, Hongyu Zhang, Wei Wu, Shaowei
  Cai, Bing He, Saravanakumar Rajmohan, et~al.
\newblock Pulns: Positive-unlabeled learning with effective negative sample
  selector.
\newblock In \emph{Proceedings of the AAAI Conference on Artificial
  Intelligence}, volume~35, pages 8784--8792, 2021.

\bibitem[Maas et~al.(2011)Maas, Daly, Pham, Huang, Ng, and Potts]{IMDB2011}
Andrew~L. Maas, Raymond~E. Daly, Peter~T. Pham, Dan Huang, Andrew~Y. Ng, and
  Christopher Potts.
\newblock Learning word vectors for sentiment analysis.
\newblock In \emph{Proceedings of the 49th Annual Meeting of the Association
  for Computational Linguistics: Human Language Technologies}, pages 142--150,
  Portland, Oregon, USA, June 2011. Association for Computational Linguistics.

\bibitem[Marcu et~al.(2019)Marcu, Bichmann, Kuchenbecker, Kowalewski,
  Freudenmann, Backert, M\"{u}hlenbruch, Szolek, L\"{u}bke, Wagner, Engler,
  Matovina, Wang, Hauri-Hohl, Martin, Kapolou, Walz, Velz, Moch, Regli,
  Silginer, Weller, L\"{o}ffler, Erhard, Schlosser, Kohlbacher,
  Stevanovi{\'{c}}, Rammensee, and Neidert]{hlala}
Ana Marcu, Leon Bichmann, Leon Kuchenbecker, Daniel~Johannes Kowalewski,
  Lena~Katharina Freudenmann, Linus Backert, Lena M\"{u}hlenbruch, Andr{\'{a}}s
  Szolek, Maren L\"{u}bke, Philipp Wagner, Tobias Engler, Sabine Matovina, Jian
  Wang, Mathias Hauri-Hohl, Roland Martin, Konstantina Kapolou, Juliane~Sarah
  Walz, Julia Velz, Holger Moch, Luca Regli, Manuela Silginer, Michael Weller,
  Markus~W. L\"{o}ffler, Florian Erhard, Andreas Schlosser, Oliver Kohlbacher,
  Stefan Stevanovi{\'{c}}, Hans-Georg Rammensee, and Marian~Christoph Neidert.
\newblock The {HLA} ligand atlas - a resource of natural {HLA} ligands
  presented on benign tissues.
\newblock September 2019.
\newblock \doi{10.1101/778944}.

\bibitem[Menon et~al.(2015)Menon, Rooyen, Ong, and
  Williamson]{Menon2015LearningFC}
Aditya Menon, Brendan~Van Rooyen, Cheng~Soon Ong, and Bob Williamson.
\newblock Learning from corrupted binary labels via class-probability
  estimation.
\newblock In Francis Bach and David Blei, editors, \emph{Proceedings of the
  32nd International Conference on Machine Learning}, volume~37 of
  \emph{Proceedings of Machine Learning Research}, pages 125--134, Lille,
  France, 07--09 Jul 2015. PMLR.

\bibitem[M\"{u}ller et~al.(2019)M\"{u}ller, Kornblith, and Hinton]{labelsmooth}
Rafael M\"{u}ller, Simon Kornblith, and Geoffrey~E Hinton.
\newblock When does label smoothing help?
\newblock In H.~Wallach, H.~Larochelle, A.~Beygelzimer, F.~d\textquotesingle
  Alch\'{e}-Buc, E.~Fox, and R.~Garnett, editors, \emph{Advances in Neural
  Information Processing Systems}, volume~32, 2019.

\bibitem[Nielsen et~al.(2005)Nielsen, Lundegaard, Lund, and
  Ke{\c{s}}mir]{Nielsen_2005}
Morten Nielsen, Claus Lundegaard, Ole Lund, and Can Ke{\c{s}}mir.
\newblock The role of the proteasome in generating cytotoxic t-cell epitopes:
  insights obtained from improved predictions of proteasomal cleavage.
\newblock \emph{Immunogenetics}, 57\penalty0 (1-2):\penalty0 33--41, mar 2005.
\newblock \doi{10.1007/s00251-005-0781-7}.
\newblock URL \url{https://doi.org/10.1007%2Fs00251-005-0781-7}.

\bibitem[Ovadia et~al.(2019)Ovadia, Fertig, Ren, Nado, Sculley, Nowozin,
  Dillon, Lakshminarayanan, and Snoek]{ovadia2019can}
Yaniv Ovadia, Emily Fertig, Jie Ren, Zachary Nado, David Sculley, Sebastian
  Nowozin, Joshua Dillon, Balaji Lakshminarayanan, and Jasper Snoek.
\newblock Can you trust your model's uncertainty? evaluating predictive
  uncertainty under dataset shift.
\newblock \emph{Advances in neural information processing systems}, 32, 2019.

\bibitem[Purcell et~al.(2019)Purcell, Ramarathinam, and Ternette]{mass_spec}
Anthony~W. Purcell, Sri~H. Ramarathinam, and Nicola Ternette.
\newblock Mass spectrometry{\textendash}based identification of {MHC}-bound
  peptides for immunopeptidomics.
\newblock \emph{Nat Protoc}, 14\penalty0 (6):\penalty0 1687--1707, may 2019.
\newblock \doi{10.1038/s41596-019-0133-y}.

\bibitem[Rizve et~al.(2021)Rizve, Duarte, Rawat, and Shah]{Rizve2021}
M.~N. Rizve, Kevin Duarte, Y.~Rawat, and M.~Shah.
\newblock In defense of pseudo-labeling: An uncertainty-aware pseudo-label
  selection framework for semi-supervised learning.
\newblock \emph{International Conference on Learning Representations}, 2021.

\bibitem[Ruff et~al.(2018)Ruff, Vandermeulen, Goernitz, Deecke, Siddiqui,
  Binder, M{\"u}ller, and Kloft]{ruff2018deep}
Lukas Ruff, Robert Vandermeulen, Nico Goernitz, Lucas Deecke, Shoaib~Ahmed
  Siddiqui, Alexander Binder, Emmanuel M{\"u}ller, and Marius Kloft.
\newblock Deep one-class classification.
\newblock In \emph{International conference on machine learning}, pages
  4393--4402. PMLR, 2018.

\bibitem[Schatz et~al.(2008)Schatz, Peters, Akkad, Ullrich, Martinez, Carroll,
  Bulik, Rammensee, van Endert, Holzh\"{u}tter, Tenzer, and Schild]{Schatz2008}
Mark~M. Schatz, Bj\"{o}rn Peters, Nadja Akkad, Nina Ullrich, Alejandra~Nacarino
  Martinez, Oliver Carroll, Sascha Bulik, Hans-Georg Rammensee, Peter van
  Endert, Hermann-Georg Holzh\"{u}tter, Stefan Tenzer, and Hansj\"{o}rg Schild.
\newblock Characterizing the n-terminal processing motif of {MHC} class i
  ligands.
\newblock \emph{The Journal of Immunology}, 180\penalty0 (5):\penalty0
  3210--3217, February 2008.
\newblock \doi{10.4049/jimmunol.180.5.3210}.
\newblock URL \url{https://doi.org/10.4049/jimmunol.180.5.3210}.

\bibitem[Shi et~al.(2018)Shi, Gong, Ding, Tao, and Zheng]{Shi2018}
Weiwei Shi, Yihong Gong, Chris Ding, Zhiheng~MaXiaoyu Tao, and Nanning Zheng.
\newblock Transductive semi-supervised deep learning using min-max features.
\newblock In \emph{Proceedings of the European Conference on Computer Vision
  (ECCV)}, pages 299--315, 2018.

\bibitem[Su et~al.(2021)Su, Chen, and Xu]{imbnnpu}
Guangxin Su, Weitong Chen, and Miao Xu.
\newblock Positive-unlabeled learning from imbalanced data.
\newblock In \emph{Proceedings of the Thirtieth International Joint Conference
  on Artificial Intelligence}. International Joint Conferences on Artificial
  Intelligence Organization, aug 2021.
\newblock \doi{10.24963/ijcai.2021/412}.

\bibitem[Tanaka et~al.(2018)Tanaka, Ikami, Yamasaki, and Aizawa]{Tanaka2018}
Daiki Tanaka, Daiki Ikami, Toshihiko Yamasaki, and Kiyoharu Aizawa.
\newblock Joint optimization framework for learning with noisy labels.
\newblock In \emph{Proceedings of the IEEE Conference on Computer Vision and
  Pattern Recognition}, pages 5552--5560, 2018.

\bibitem[Toussaint et~al.(2011)Toussaint, Maman, Kohlbacher, and
  Louzoun]{toussaint_universal_2011}
Nora~C. Toussaint, Yaakov Maman, Oliver Kohlbacher, and Yoram Louzoun.
\newblock Universal peptide vaccines – {Optimal} peptide vaccine design based
  on viral sequence conservation.
\newblock \emph{Vaccine}, 29\penalty0 (47):\penalty0 8745--8753, November 2011.
\newblock ISSN 0264410X.
\newblock \doi{10.1016/j.vaccine.2011.07.132}.

\bibitem[Van~Engelen and Hoos(2020)]{VanEngelen2020}
Jesper~E Van~Engelen and Holger~H Hoos.
\newblock A survey on semi-supervised learning.
\newblock \emph{Machine Learning}, 109\penalty0 (2):\penalty0 373--440, 2020.

\bibitem[Vita et~al.(2018)Vita, Mahajan, Overton, Dhanda, Martini, Cantrell,
  Wheeler, Sette, and Peters]{iedb}
Randi Vita, Swapnil Mahajan, James~A Overton, Sandeep~Kumar Dhanda, Sheridan
  Martini, Jason~R Cantrell, Daniel~K Wheeler, Alessandro Sette, and Bjoern
  Peters.
\newblock {The Immune Epitope Database (IEDB): 2018 update}.
\newblock \emph{Nucleic Acids Research}, 47\penalty0 (D1):\penalty0 D339--D343,
  10 2018.
\newblock ISSN 0305-1048.
\newblock \doi{10.1093/nar/gky1006}.

\bibitem[Xiao et~al.(2017)Xiao, Rasul, and Vollgraf]{xiao2017fashion}
Han Xiao, Kashif Rasul, and Roland Vollgraf.
\newblock Fashion-mnist: a novel image dataset for benchmarking machine
  learning algorithms.
\newblock \emph{arXiv preprint arXiv:1708.07747}, 2017.

\bibitem[Xu et~al.(2017)Xu, Xu, Xu, and Tao]{xu2017multi}
Yixing Xu, Chang Xu, Chao Xu, and Dacheng Tao.
\newblock Multi-positive and unlabeled learning.
\newblock In \emph{Proceedings of the Twenty-Sixth International Joint
  Conference on Artificial Intelligence}, pages 3182--3188, 2017.

\bibitem[Zhang et~al.(2018)Zhang, Cisse, Dauphin, and Lopez-Paz]{mixup}
Hongyi Zhang, Moustapha Cisse, Yann~N. Dauphin, and David Lopez-Paz.
\newblock Mixup: Beyond empirical risk minimization.
\newblock \emph{International Conference on Learning Representations}, 2018.

\bibitem[Zhao et~al.(2022)Zhao, Xu, Jiang, Wen, and Huang]{distpu}
Yunrui Zhao, Qianqian Xu, Yangbangyan Jiang, Peisong Wen, and Qingming Huang.
\newblock Dist-pu: Positive-unlabeled learning from a label distribution
  perspective.
\newblock In \emph{Proceedings of the IEEE/CVF Conference on Computer Vision
  and Pattern Recognition (CVPR)}, pages 14461--14470, June 2022.

\end{thebibliography}
}

\end{document}